\documentclass[sigconf, nonacm]{acmart}
\usepackage{amsmath,amsfonts}
\usepackage[ruled]{algorithm2e} 
\usepackage{algorithmic}
\usepackage{array}
\usepackage{textcomp}
\usepackage{stfloats}
\usepackage{url}
\usepackage{verbatim}
\usepackage{graphicx}

\usepackage{amsthm}
\usepackage{multirow}

\usepackage{color}
\usepackage{xcolor}
\usepackage{colortbl}

\newcommand\vldbdoi{XX.XX/XXX.XX}

\newcommand\vldbpagestyle{plain} 

\begin{document}

\title{SecureBoost+ : Large Scale and High-Performance Vertical Federated Gradient Boosting Decision Tree}

\author{Tao Fan$^{*\dagger}$}
\affiliation{%
  \institution{Hong Kong University of Science and Technology, China}
}

\affiliation{%
  \institution{WeBank, China}
}
\email{tfanac@cse.ust.hk}

\author{Weijing Chen$^*$}
\affiliation{%
  \institution{WeBank, China}
}
\email{weijingchen@webank.com}

\author{Guoqiang Ma}
\affiliation{%
  \institution{WeBank, China}
}
\email{zotrseeewma@webank.com}

\author{Yan Kang}
\affiliation{%
  \institution{WeBank, China}
}
\email{yangkang@webank.com}

\author{Lixin Fan}
\affiliation{%
  \institution{WeBank, China}
}
\email{lixinfan@webank.com}

\author{Qiang Yang}
\affiliation{%
  \institution{Hong Kong University of Science and Technology, China}
}

\affiliation{%
  \institution{WeBank, China}
}
\email{qyang@cse.ust.hk}

\begin{abstract}
Gradient boosting decision tree (GBDT) is an ensemble machine learning algorithm, which is widely used in industry, due to its good performance and easy interpretation. Due to the problem of data isolation and the requirement of privacy, many works try to use vertical federated learning to train machine learning models collaboratively with privacy guarantees between different data owners. SecureBoost is one of the most popular vertical federated learning algorithms for GBDT. However, in order to achieve privacy preservation, SecureBoost involves complex training procedures and time-consuming cryptography operations. This causes SecureBoost to be slow to train and does not scale to large scale data.

In this work, we propose SecureBoost+, a large-scale and high-performance vertical federated gradient boosting decision tree framework. SecureBoost+ is secure in the semi-honest model, which is the same as SecureBoost. SecureBoost+ can be scaled up to tens of millions of data samples easily. SecureBoost+ achieves high performance through several novel optimizations for SecureBoost, including ciphertext operation optimization, the introduction of new training mechanisms, and multi-classification training optimization. The experimental results show that SecureBoost+ is 6-35x faster than SecureBoost, but with the same accuracy and can be scaled up to tens of millions of data samples and thousands of feature dimensions.

\end{abstract}

\maketitle
\def\thefootnote{*}\footnotetext{These authors contributed equally to this work}\def\thefootnote{\arabic{footnote}}
\def\thefootnote{$\dagger$}\footnotetext{Corresponding author}\def\thefootnote{\arabic{footnote}}
\pagestyle{\vldbpagestyle}





\section{Introduction}
With the rapid development of information technology, a large number of valuable data have been produced in various industrial fields.  Artificial intelligence technology, especially deep learning, has made amazing progress in mining massive amounts of data. However, data is also a key constraint on the application of AI.  In a real-world scenario, data is stored in different organizations or different departments of the same enterprise, forming a phenomenon known as data isolation. With the increasing attention of the whole society to data privacy protection, most countries are developing laws and regulations related to data privacy protection. For example, the European Union has enacted the General Data Protection Regulation (GDPR)~\cite{voigt2017eu}, which aims to strengthen the security of user data privacy. This trend brings new challenges to cross-organizational data sharing and AI applications.

Federated learning (FL)~\cite{mcmahan2017communication,yang2019federated}, a distributed collaborative machine learning paradigm, is a promising approach to deal with the data isolation challenge. It enables local models to benefit from all parties while keeping local data private \cite{kairouz2021advances}. In addition, combined with various cryptography techniques, local data privacy can be effectively protected. Federated learning can be generally divided into vertical federated learning~\cite{liu2024vfl}, horizontal federated learning, and federated transfer learning. Vertical federated learning on logistic regression~\cite{hardy2017private,chen2021homomorphic,yang2019quasi}, GBDT~\cite{cheng2021secureboost}, neural network~\cite{zhang2018gelu,zhang2020additively,fu2022blindfl,kang2022privacy,kang2022fedcvt}, transfer learning~\cite{liu2020secure}, etc, have been previously studied.

Gradient boosting decision tree (GBDT) is an ensemble machine learning algorithm that is widely used in industry~\cite{chen2016xgboost,ke2017lightgbm, dorogush2018catboost} due to its good performance and easy interpretation. For example, GBDT and its variants are usually used in recommendation~\cite{ke2017lightgbm,shahbazi2019product}, fraud detection~\cite{cao2019titant}, and click-through rate prediction~\cite{wang2018tem} tasks with large-scale data. Due to the advantage of GBDT, many works try to apply vertical federated learning to GBDT model~\cite{cheng2021secureboost,fu2021vf2boost,li2020practical}. Our previous work, SecureBoost~\cite{cheng2021secureboost}, is one of the most popular vertical federated learning algorithms for GBDT. However, as far as we know, some previous studies have focused more on the security of algorithms and less on the gap between research and industrial applications. Some works are only tested on public experimental data sets instead of million-scale or high-dimensional data, which are common in real-world scenarios, and some works are tested on relatively larger data sets but can not be completed in a reasonable time. And according to our practical experiences, the performance of SecureBoost is still unsatisfactory.

In this work, we propose SecureBoost+, a large-scale and high-performance vertical federated gradient boosting decision tree framework. It is developed based on the work SecureBoost~\cite{cheng2021secureboost}, which has been integrated to FATE~\cite{JMLR:v22:20-815}. The main contributions of SecureBoost+ are:

\begin{itemize}
\item We propose a ciphertext operation optimization framework based on vertical federated gradient boosting decision tree methods, designed for Homomorphic Encryption(HE) \cite{paillier1999public} encryption schema.

\item We propose two novel training mechanisms for SecureBoost+: Mix Tree mode and Layered Tree mode. The novel training mechanisms can significantly reduce the number of interactions and communication costs between parties.

\item We propose a new multi-classification mechanism named SecureBoost-MO to speed up multi-classification training.

\item We investigate the performance of SecureBoost+ on a large-instance and a high-dimension dataset. Experimental results demonstrate that SecureBoost+ is 6-35x faster than SecureBoost, but with the same accuracy, and can be scaled up to tens of millions of data samples and thousands of feature.

\end{itemize}

\section{PRELIMINARIES}

\subsection{Gradient Boosting Decision Tree}
GBDT is an ensemble machine learning algorithm that is widely used in industry\cite{chen2016xgboost,ke2017lightgbm}. Let ${f}$ be a decision tree function, the prediction for an instance is given by the sum of all $K$ decision tree is obtained by:
\begin{equation}
    \hat{y_i} = \sum_{k=1}^{K} f_{k}(\textbf{x}_i) 
\end{equation}

When training an ensemble tree model, we add a new decision tree $f_t$ at iteration $t$ to minimize the following second-order approximation loss:
\begin{equation}
    L^{(t)} = \sum_{i \in \textbf{I}} \left [  l(y_i, \hat{y}^{(t-1)}_{i}) + g_i f_t(\textbf{x}_i) +
   \frac{1}{2} h_i f_{t}^{2}(\textbf{x}_i) \right ] + \Omega(f_t)
   \label{eq:gbdtloss}
\end{equation}

where $\Omega(f_t)$ is the regularization term and $g_i$, $h_i$ are the first and second derivatives of $l(y_i, \hat{y}^{(t-1)}_{i})$ with respect to $\hat{y}_i$.

We rewrite equation \ref{eq:gbdtloss} in a leaf-weight format~\cite{chen2016xgboost}:
\begin{equation}
    L^{(t)} = \sum_{j \in \textbf{T}}\sum_{i \in \textbf{I}_j} \left [  l(y_i, \hat{y}^{(t-1)}_{i}) + g_i w_j +
   \frac{1}{2} h_i w_{j}^{2}\right ] + \frac{\lambda}{2}w_{j}^{2}
   \label{eq:gbdtloss2}
\end{equation}

By setting the second-order approximation function above, we derive the split gain function for splitting a node:
\begin{equation}
    gain = \frac{1}{2} \left [ \frac{ (\Sigma_{i \in I_L} g_{i})^{2} }{\Sigma_{i \in I_L} h_i + \lambda} + 
    \frac{ (\Sigma_{i \in I_R} g_{i})^{2} }{\Sigma_{i \in I_R} h_i + \lambda} - 
    \frac{ (\Sigma_{i \in I} g_{i})^{2} }{\Sigma_{i \in I} h_i + \lambda} \right ]
\end{equation}
where $I_L$, $I_R$, and $I$ are the instances space of the left, right, and parent nodes, respectively. The leaf output of an arbitrary leaf $j$ put is given by:
\begin{equation}
    w_j = -\frac{\Sigma_{i \in I_j} g_j}{\Sigma_{i \in I_j} h_j + \lambda}
\end{equation}  

\subsection{Paillier Homomorphic Encryption}
The Paillier Homomorphic Encryption (PHE)\cite{paillier1999public} schema is an additive homomorphic cryptosystem, and it is the key component of many privacy-preserving ML algorithms\cite{cheng2021secureboost,chai2020secure,yang2019federated}.
The core properties of PHE are homomorphic addition and scalar multiplication:
\begin{equation}
\label{he_add}
    [[x_{1}]] \oplus [[x_{2}]] = [[x_{1} + x_{2}]]
\end{equation}
\begin{equation}
\label{he_mulply}
    x_{1} \otimes [[x_{2}]] = [[x_{1} \times x_{2}]]
\end{equation}
where $[[.]]$ denotes the encryption operator.

\subsection{SecureBoost}

In SecureBoost \cite{cheng2021secureboost}, all parties conduct a privacy-preserving instance intersection at the beginning. After the intersection, we get $\left \{  \textbf{X}^{k} \in \mathbb{R}^{n \times  d_k} \right \}_{k=1}^{m}$ as the final aligned data matrices distributed on each party. Each party holds a data matrix $\textbf{X}^{d_k}$ with $n$ instances and $d_k$ features. For ease of description, we use $\textbf{X}$ to denote a federated instance whose features are distributed on multiple parties and use $\textbf{I} = \{1, ... n\}$ to represent the instance space. $\textbf{F}^{k} \in \left \{ f_1, f_2, ..., f_{d_k}\right \}$ is the feature set on the $k$-th party. 
Each party holds its unique features, for any two feature sets $F^i$ and $F^j$ from $i$-th party and $j$-th party, $\textbf{F}^i \cap \textbf{F}^j = \emptyset$. We define two party roles: $\textbf{Active Party}$ and $\textbf{Passive Party}$. The active party is the data provider who holds both a data matrix and the class label $\textbf{y} \in \mathbb{R}^n$.
The passive party is the data provider who only holds a data matrix and does not hold the class label.

The goal of SecureBoost is to train an ensemble gradient boosting tree model on the federated dataset $\left \{  \textbf{X}^{k} \in \mathbb{R}^{n \times  d_k} \right \}_{k=1}^{m}$ and label $\textbf{y}$. SecureBoost uses a histogram-based split finding strategy similar to XGBoost \cite{chen2016xgboost}. Therefore, before constructing the tree,  every party uses the quantile binning method to transform their feature values into bin indices.

To ensure that the label information is not leaked through from $g_i$ and $h_i$, the active party needs to apply homomorphic encryption on $g_i$ and $h_i$ and then send them to the passive party. With the additive property of homomorphic encryption on $g_i$ and $h_i$, the passive party can calculate the ciphertext histogram and then construct split-info. To protect the feature information of the passive party, the passive party marks split-info with unique IDs, randomly shuffles this split-info, and then sends it to the active party. The global optimal split node using features of all parties can be calculated collaboratively through the above process. The active party can calculate the local split information and then receive the encrypted, split information from the passive party and decrypt it. The active party is able to find the optimal split node without leaking labels and knowing information about the passive party features. The party that holds the optimal split information will be responsible for splitting and assigning current node instances to child nodes. The result of instances assignment will be synchronized to all parties. Starting at the root node, We repeat the split finding layer-wise until the convergence condition is reached.

\subsection{Performance Bottlenecks Analysis for SecureBoost}

The most expensive parts of histogram-based gradient boosting algorithms are the histogram building and split node finding~\cite{ke2017lightgbm,chen2016xgboost}. Due to homomorphic encryption operations, these two parts are particularly expensive to compute in SecureBoost.  From the review of SecureBoost, we have the following observations:

\begin{itemize}
 \item  \textbf{Encryption and Addition Operation Cost}: Histogram calculation of the passive party will be especially time-consuming because it is calculated on the homomorphic ciphertext. Homomorphic addition to ciphertext usually involves multiplication and modular operations on large integers and is much more expensive than plaintext addition~\cite{paillier1999public}.

 \item  \textbf{Decryption Operation Cost}: Since we know that the gradient and hessian are encrypted, we have to decrypt a batch of encrypted split information when we split the node. Decryption is expensive to train on large amounts of data.

 \item \textbf{Communication Cost}: At the beginning of building each tree, we need to synchronize the encryption gradient and hessian to the passive parties. At the same time, a batch of encrypted split information needs to be sent to calculate the split node of each layer of the tree. This entails heavy communication overhead.

\end{itemize}

\section{Proposed SecureBoost+ Framework}
In this section, we propose our solution to improve the performance of SecureBoost in the following directions.

\begin{itemize}
\item \textbf{Ciphertext Operation Optimization}: 
In order to reduce the computation cost of homomorphic ciphertext on passive parties, especially the cost of histogram computation and split node finding. We propose a ciphertext operation optimization framework. Main idea: Pack Plaintext reduces the amount of ciphertext in computation and transmission. The ciphertext of the passive parties is compressed to reduce decryption and communication costs.

\item \textbf{Training Mechanism Optimization}:  At the level of training mechanism, two optimization methods of mix tree and layered tree are designed to reduce the number of interactions and communication cost between parties. 

\item \textbf{SecureBoost-MO Optimization}: We propose a new multi-classification mechanism named SecureBoost-MO to speed up multi-classification training.

\item \textbf{Other Optimization}: We integrate common engineering optimization techniques into SecureBoost+. They directly reduce instances and data terms involved in boosting tree training. This is very useful when training on large-scale data. 

\end{itemize}

\subsection{Ciphertext Operation Optimization}
\subsubsection{Rough Estimate of Ciphertext Operation Cost}
Suppose we train a binary-classification task, we have $n_{i}$ samples, $n_{f}$ features, every feature has $n_{b}$ bins, the tree depth is $d$, and then there are $n_{n}=2^{d}$ nodes.
We have a rough cost estimate for building a decision tree in SecureBoost:
\begin{itemize}
    \item The ciphertext computation cost: ciphertext histogram computation + bin cumsum:
    \begin{equation}
        cost_{comp} = 2 \times n_{i} \times d + 2 \times n_{n} \times n_f \times n_b
        \label{comp}
    \end{equation}
    \item The encryption and decryption cost: $g\&h$ encryption + split-info decryption:
    \begin{equation}
        cost_{ende} = 2 \times n_{i} + 2 \times n_{b} \times n_f \times n_n
        \label{ende}
    \end{equation}
    \item The communication cost: encrypted $g\&h$ + split-infos batches:
    \begin{equation}
        cost_{comm} = 2 \times n_i + 2 \times n_b \times n_f \times n_n
        \label{comm}
    \end{equation}
\end{itemize}

\subsubsection{GH Packing}
We first introduce GH Packing. From SecureBoost,  we notice that gradient and hessian have the same operations in the process of histogram construction and split-finding: the gradient and hessian of a sample will be encrypted at the same time, and the encrypted gradient and hessian will go to the same bucket, but be added to different indices. The gradient and hessian of a split-info will also be decrypted together.
 
The Paillier\cite{paillier1999public} is a homomorphic encryption algorithm used in SecureBoost. During training, a key length of 1024 or 2048 is usually used. When 1024 key length is used, the upper bound on the positive integer allowed in encryption is usually 1023 bits in length. This upper bound is much larger than the gradient and hessian at fixed points. As a result, a lot of plaintext space is wasted in the encryption process. Inspired by recently proposed gradient quantization and gradient batching techniques~\cite{zhang2020batchcrypt}, we adopt a GH(gradient: $g$ and hessian: $h$) Packing method in our work. We bundle $g$ and $h$ using bit arithmetic: We move g to the left by a certain number of bits and then add h to it.

For example, in a binary-classification task, the range of $g$ is $[-1, 1]$,  $g_{max} = 1$ and the range of $h$ is $[0, 1]$, $h_{max} = 1$. Considering that g might be negative, we can offset g to ensure that it is a positive number. In the classification case, the offset number $g_{off}$ for $g$ will be $1$. In the split finding step, we can remove the offset number when recovering $g,h$.

After offsetting $g$, we adopt a fixed-point encoding strategy to encode $g$ and $h$, which is shown in equation \ref{gh_all}, where $r$ is usually 53. Then, we can pack $g$ and $h$ into one large positive number. To avoid overflowing aggregation results during histogram calculation, we need to reserve more bits for $g$ and $h$. Assuming we have $n_i$ instances, we need to make sure that no cumulative gradients and hessians in the histogram will be larger than $g_{cmax}$ and $h_{cmax}$ in equation \ref{gh_cum_max}. Therefore, we assign $b_{g}$ bits for $g$ and $b_{h}$ bits for $h$ in equation \ref{gh_cum_max}:
\begin{equation}
\begin{aligned}
    g_{cmax} &= n_i \times \lfloor (g_{max} + g_{off}) \times 2^{r} \rfloor\\
    h_{cmax} &= n_i \times \lfloor h_{max} \times 2^{r} \rfloor\\
    b_{g} &= BitLength(g_{cmax}) \\
    b_{h} &= BitLength(h_{cmax}) \\
    b_{gh} &= b_{g} + b_{h}
    \label{gh_cum_max}
\end{aligned}
\end{equation}

Finally, we calculate a large positive number $gh$ and homomorphic encryption ciphertext $[[gh]]$ as follow:
\begin{equation}
\begin{aligned}
    g &= \lfloor(g + g_{off}) \times 2^{r} \rfloor\\
    h &= \lfloor h \times 2^{r} \rfloor\\
    gh &= g<< b_h + h \\
    [[gh]] &= HE(gh)
     \label{gh_all}
\end{aligned}
\end{equation}
  
\begin{figure}[ht!]
    \centering
\includegraphics[width=0.48\textwidth]{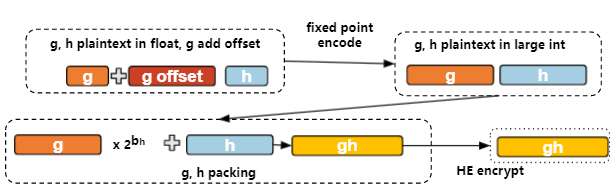}
    \caption{The Process of GH Packing}
    \label{ghpacking}
\end{figure}

Figure \ref{ghpacking} shows details of the GH packing process. The active party synchronizes $[[gh]]$ ciphertext to the passive party. The passive party only needs to calculate one ciphertext per sample when constructing the histogram, while in SecureBoost, it needs to calculate two ciphertexts per sample. From the GH Packing optimization, the whole ciphertext computation cost is \textbf{reduced by half}. 


\subsubsection{Ciphertext Compression}
Although the GH packing algorithm uses a lot of plaintext space, the plaintext space is not fully utilized. Remind that we usually have a 1023 bit-length integer as the plaintext upper bound when using a 1024-bit key length in the Paillier \cite{paillier1999public}. Assuming instances number $n$ is $1,000,000$, $r$ is $53$, we will assign $74$ bits for $g$ and $73$ bits for $h$, according to equation \ref{gh_cum_max}. The $b_{gh}$ is 147, which is still less than 1023.
 
To fully use the plaintext space, we can perform ciphertext compression on the passive parties.  This technique utilizes the property of our built-in HE algorithm, which is that an addition operation or a scalar multiplication operation is less costly than a decryption operation.  We use addition and multiplication operations to compress several encrypted numbers in split-info into one encrypted number and make sure that the plaintext space is fully utilized. The idea of this method is the same as the GH Packing method: we multiply a ciphertext $C$ by $2^{b_{gh}}$, shifting its plaintext by ${b_{gh}}$ bits to the left. Then, we can add another ciphertext to it, compressing the two ciphertexts into one. We can repeat this process until no more bits are left in the plaintext space to compress, as shown in equation \ref{compress}. Once the active party receives an encrypted number, it only needs to decrypt one encrypted number to recover several $gh$ statistical results and reconstruct split-info.

\begin{equation}
\begin{aligned}
    C_{cum0} &= C_{0} \\
    C_{cum1} &= C_{cum0} \times 2^{b_{gh}} + C_{1} \\
    .... \\
    C_{cumk} &= C_{cumk-1} \times 2^{b_{gh}} + C_{k-1}
    \label{compress}
\end{aligned}
\end{equation}

\begin{figure*}[htbp]
    \centering
    \includegraphics[width=0.90\textwidth]{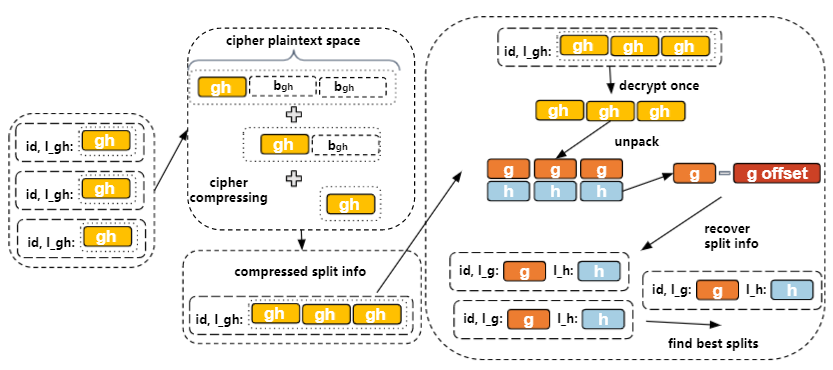}
    \caption{The Process of Ciphertext Compression.}
    \label{cipher-compress}
\end{figure*}

In the case above, given a 1023 bit-length plaintext space and $b_{gh}$ of 147, We are able to compress $\lfloor 1023 / 147 \rfloor= 6$ split points into 1.  By using ciphertext compression, decryption operations and decryption time can be reduced by up to 6 times. On the other hand, the communication cost of encrypted split-info is also reduced by up to 6 times. Figure \ref{cipher-compress} illustrates the process of ciphertext compression.


\subsubsection{Ciphertext Histogram Subtraction}
Histogram subtraction is a classic optimization technique in the histogram-based tree algorithms. When splitting nodes, samples are distributed to either the left or right nodes. The parent histogram is obtained for each feature by adding the set of histograms of the left and right children nodes. To take advantage of this property, in SecureBoost+, both the active and passive parties cache the histogram of the previous layer.  When growing from the current layer, the active and passive parties first calculate the histogram of nodes with fewer samples and finally obtain the histogram of sibling nodes by histogram subtraction with the previous layer cache information. Since the ciphertext histogram calculation for passive parties is always the largest overhead, the ciphertext histogram subtraction will greatly speed up the calculation, and at least half of the ciphertext calculation is reduced.

\subsubsection{Re-estimate Cost after Ciphertext Operation Optimization}
Let's go back to cost estimates. In more detail, suppose we use paillier encryption algorithm with 1024 key length, we have 1 million instances and 2,000 features, tree depth is 5, bin number is 32, the precision parameter $r$ of equation \ref{gh_all}  is 53.  

With GH packing and histogram subtraction optimization, ciphertext histogram computation cost is reduced by 4 times and operations in bin cumsum is reduced by half. 

 The ciphertext computation cost is calcualted by: 
 \begin{equation}
    cost_{comp}^{*} = \frac{1}{2} \times n_i \times h \times n_f + n_n \times n_f \times n_b
    \label{comp-opt}
 \end{equation}

We compare the values of the equation (\ref{comp}) and (\ref{comp-opt}), the ciphertext computation cost is reduced by $\textbf{75\%}$.

Because we pack gradients $g$ and $h$ together, the encryption/decryption operation is reduced by half. Under current paillier encryption setting, we can compress $\eta_{s} = [1023 \div 147]=6$ ciphertexts into one ciphertext, then we get:
\begin{equation}
    cost_{ende}^{*} =  n_i + \frac{n_b \times n_f \times n_n }{\eta_{s}}
    \label{ende-opt}
\end{equation}

The communication cost is:
\begin{equation}
    cost_{comm}^{*} =  n_i + \frac{n_b \times n_f \times n_n }{\eta_{s}}
    \label{comm-opt}
\end{equation}

We compare the values of the equations (\ref{ende-opt}), (\ref{comm-opt}), and the equations (\ref{ende}), (\ref{comm}),  the cost is reduced by \textbf{78\%}.

From the estimation, we can see that ciphertext operation optimization is theoretically effective. 

\subsection{Training Mechanism Optimization}
At the training mechanism level, two optimization methods are proposed: mixed tree mode and layered tree mode. These two methods can significantly reduce the number of interactions and communication cost between parties.

\subsubsection{Mix Tree Mode}

In the Mix Tree Mode, every party will build a certain number of trees using their local features, and this procedure repeats until the max epoch is reached or the stop condition is met.

\begin{figure}[!ht]
    \includegraphics[width=3.3in]{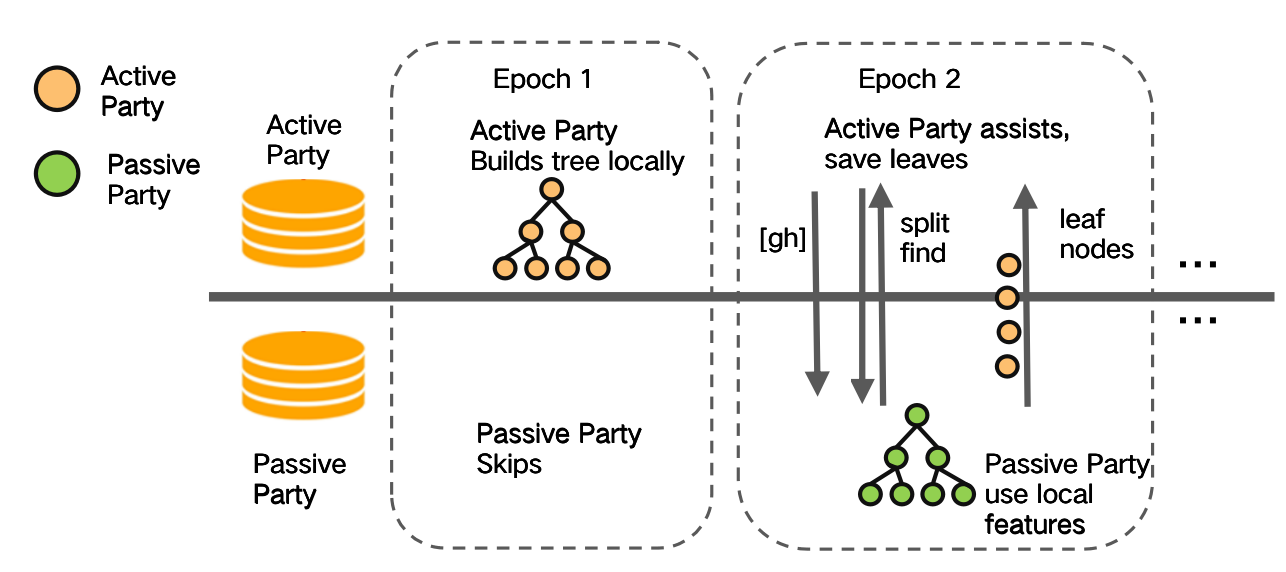}
    \caption{The Mix Tree Mode.}
    \label{mix-tree}
\end{figure}

When the active party builds a decision tree, the active party simply builds the tree locally. When the tree is built in the passive party, the relevant passive party receives the encrypted g/h, assisted by the active party, to find the best split point. The structure of the tree and split point is retained on the passive party, while the leaf weight is retained on the active party. This mode is suitable for vertical federated learning under data balancing. In this way, the number of interactions and communication costs can be significantly reduced, thus speeding up training. Figure \ref{mix-tree} demonstrates the mix tree training process. 

\subsubsection{Layered Tree Mode}

The Layered Tree Mode is another optimization mechanism, which is designed for vertical federated learning under data balancing.

\begin{figure}[!ht]
    \includegraphics[width=3.3in]{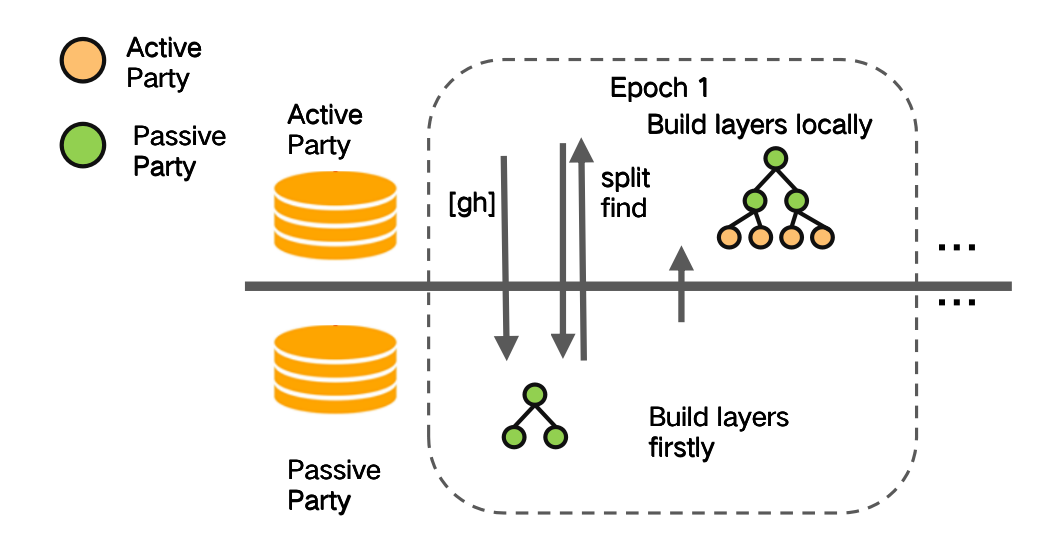}
    \caption{The Layered Tree Mode.}
    \label{layered-tree}
\end{figure}

Similar to The Mix Tree Mode, the active and the passive party are responsible for different layers in the tree when building a decision tree. Every decision tree has in total $h=h_{active}+h_{passive}$ layers.
The passive party will be responsible for building the first $h_{passive}$ layers, assisted by the active party. The active party will build the next $h_{active}$ layers. All trees will be built in this 'layered' manner. In this way, the number of interactions and communication costs can be significantly reduced, thus speeding up training. Figure \ref{layered-tree} demonstrates the layered tree training process.

      


\subsection{SecureBoost-MO Optimization}

In the traditional GBDT algorithm, the multi-classification task learning strategy is that each tree only learns one category separately. However, in the vertical federated learning scenario, this single-output multi-classification strategy has an obvious performance bottleneck: the computing cost and communication cost will increase significantly with the increase of the number of label categories.

\begin{figure}[htbp!]
    \includegraphics[width=3.3in]{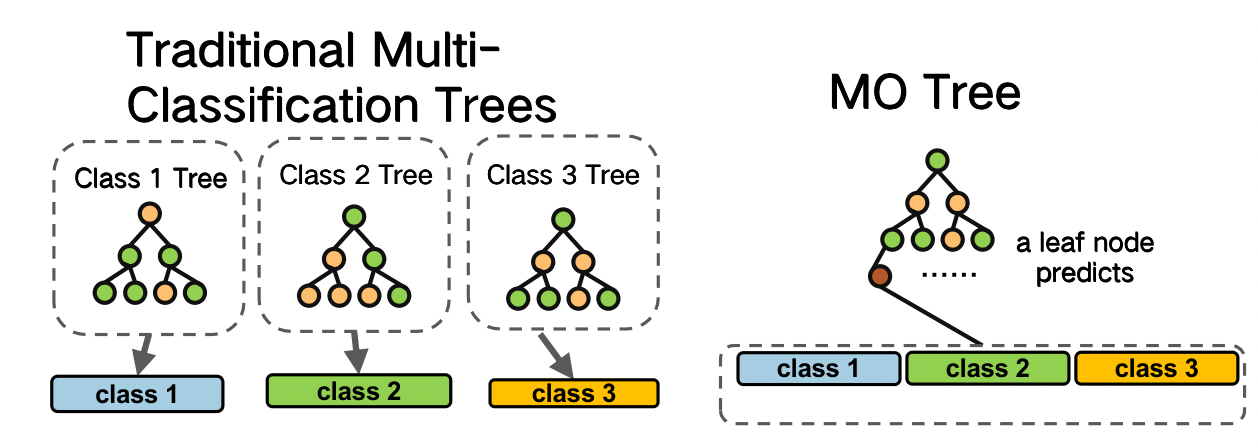}
    \caption{Traditional Trees and Multi-Output Trees.}
    \label{mo-tree-pic}
\end{figure}

Therefore, it is necessary to optimize the multi-classification task training in SecureBoost. Inspired by~\cite{zhang2020gbdt}, which introduces a general framework of multi-output gradient boosting tree, we propose a novel multi-output vertical federated boosting decision tree for multi-classification task. Leaves of multi-output tree give multi-dimension output, corresponding to every class. Instead of learning trees for every class separately,  we only need to learn one tree at each epoch. Figure \ref{mo-tree-pic} tells the differences between traditional multi-classification trees and MO-trees.
 
\subsubsection{Gain calculation of SecureBoost-MO}

We propose a split gain function for multi-output decision tree. Given a $l$ label multi-classification task, we use $\textbf{g}$ and $\textbf{h}$ to denote the gradient/hessian vectors which have $l$ elements, each element corresponds to the $g$/$h$ of 
each label. In a multi-output decision tree, every leaf obtains an output $\textbf{w}$, which also has $l$ elements. We rewrite the equation \ref{eq:gbdtloss2}:
\begin{equation}
   \begin{aligned}
           L^{(t)} = &\sum_{j \in \textbf{T}}\sum_{i \in
            \textbf{I}_j} \left [  l(\textbf{y}_i, \hat{\textbf{y}}^{(t-1)}_{i}) + \textbf{g}_i^{T} \textbf{w}_j +
        \frac{1}{2} \textbf{w}_{j}^{T} \textbf{H}_i \textbf{w}_{j}\right ] \\ &+ \frac{\lambda}{2}           \textbf{w}_{j}^{T} \textbf{w}_{j}
   \end{aligned}
   \label{eq:gbdtmoloss}
\end{equation}
    
where $\textbf{y}_i, \hat{\textbf{y}}^{(t-1)}_{i}$ are ont-hot label vector and predict score vector, $\textbf{H}$ is the hessian matrix. In SecureBoost-MO, we use cross-entropy as the multi-classification loss function, so $\textbf{H}$ is a diagonal matrix. By minimizing $L^{(t)}$, we get
\begin{equation}
    \textbf{w}_{j} = - \frac{\sum_{i \in I_j} \textbf{g}_{i} }{\sum_{i \in I_j} \textbf{h}_{i} + \lambda}
    \label{moweight}
\end{equation}

as the weight function and the split gain function:
\begin{equation}
    Score = - \frac{1}{2} \sum_{j=1}^{l} \frac{\sum_{i \in \textbf{I}}(\textbf{g}_{j})^{2}_{i}}{\sum_{i \in \textbf{I}} (\textbf{h}_{j})_{i} + \lambda}
    \label{moscore}
\end{equation}

\begin{equation}
     gain = Score_{parent}- (Score_{left} + Score_{right})  
     \label{mogain}
\end{equation}

SecureBoost-MO is based on the above leaf weight and split gain function.

\subsubsection{Split Finding of SecureBoost-MO}

The splitting finding process of Multi-output boosting tree has basically the same as that of the standard GBDT trees, but there are two differences:

\begin{itemize}
\item  In standard GBDT trees bin values are scalar $g$, $h$ values. When calculating the histogram in MO-trees, aggregated terms in the bin are \textbf{g}, \textbf{h} vectors. 

\item When calculating the split gain and leaf weights, we use mo-gain, mo-weight equation (\ref{moweight}), (\ref{mogain}). 

\end{itemize}

As a result, in the vertical federated learning scenario, we need to design a new structure that supports encryption, decryption of \textbf{g}, \textbf{h} vectors, and support operations in the histogram calculation, histogram subtraction, and split-info building. 

To fulfill this requirement, we design a Multi-Class packing algorithms based on our ciphertext optimization framework. Remind that $b_{gh}$ represent the plaintext bit length for a pair of $g$, $h$. For a HE schema that has a plaintext space of $\iota$ bits, and we have $k$ classes in total, we can pack:
    \begin{equation}
        \eta_{c} = \left \lfloor \iota \div b_{gh} \right \rfloor
    \end{equation}
    
classes in a ciphertext, and in total we need:
    \begin{equation}
        n_{k} = \left \lceil k \div  \eta_{c} \right \rceil
    \end{equation}
    
integers to pack \textbf{gh} vectors of an instance. With this strategy, we are able to pack \textbf{g} and \textbf{h} vectors of a sample into a small amount of integers and then we encrypt them to yiled a ciphertext vector. Details of the above process are shown in Algorithm \ref{algo:multi-ghpack}.

\begin{algorithm}[!ht] 
    \caption{Multi Class GH Packing}
     \label{algo:multi-ghpack}
     
    \textbf{Input:} \\
     $\textbf{G},\textbf{H}$, each row are $\textbf{g}$, $\textbf{h}$, corresponds to each instances; \\
     $n$, the instances number; \\
     $r$, the fix-point parameter;

     \textbf{Output:} \\
    $g_{off} = abs(min(\textbf{G}))$, $\textbf{G} = \textbf{G} + g_{off}$
    
    $\textbf{G}, \textbf{H} = round(\textbf{G} \times 2^{r}), round(\textbf{G} \times 2^{r})$
    
    $g_{max} = max(\textbf{G})$
    
    compute $\eta_{c}$, $b_{gh}$, $b_{g}$, $b_{h}$
    
    initialize empty matrix $[[\textbf{GH}]]$
    
    \begin{algorithmic}[1]
    \FOR{$\textbf{g}_i, \textbf{h}_i \in \textbf{G}, \textbf{H}$}
        
        \STATE init $count, e = 0, 0$
        
        \STATE init empty list $[[\textbf{vec}]]$
        
        \FOR{$g_j, h_j \in \textbf{g}_i, \textbf{h}_i$}
                
                \IF{$count == \eta_{c}$}
                    
                   \STATE $[[\textbf{vec}]].add(encrypt(e))$
                    
                   \STATE $count, e = 0, 0$
                \ENDIF
            
               \STATE $g_j = g_j << b_{h}, gh_j$ = $g_j$ + $h_j$
                
               \STATE $e << b_{gh}, e += gh_j$
                
              \STATE  $count += 1$
        \ENDFOR
        
        \STATE $[[\textbf{GH}]].add([[\textbf{vec}]])$

    \ENDFOR

    return $[[\textbf{GH}]]$
    
\end{algorithmic}   

\end{algorithm}

\begin{figure}[htbp!]
    \centering
    \includegraphics[width=3.2in]{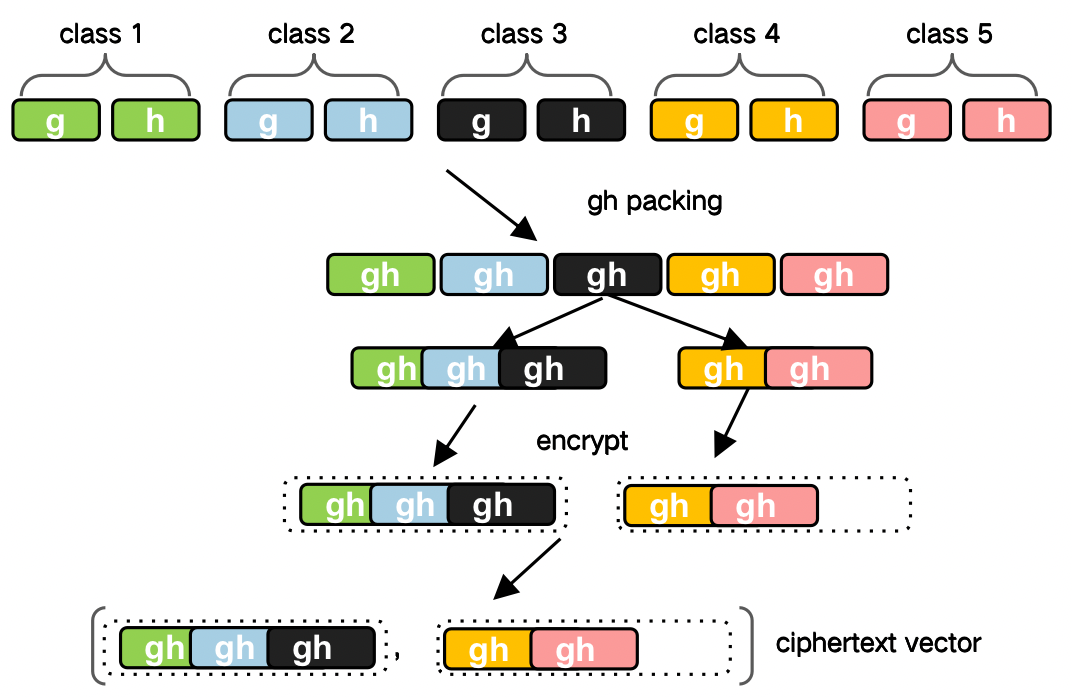}
    \caption{The Illustartion of Multi Class GH Packing}
    \label{mo-cipher-vec}
\end{figure}

Figure \ref{mo-cipher-vec} gives an intuitive illustration of Multi-Class GH Packing.

Following, we can describe the split-finding procedure of SecureBoost-MO:

\begin{itemize}

\item  At the beginning, the active party calculates the multi-classification gradient and hessian vectors for each instances. The active party packs and encrypts them using Algorithm \ref{algo:multi-ghpack}, and gets a matrix \textbf{[GH]} which contains encrypted ciphertext vectors.

\item The active party sends \textbf{[GH]} to all the passive parties. The passive parties calculate histograms and construct split-info directly on ciphertext vectors. 

\item The active party decrypts split-info and recovers accumulated \textbf{g}, \textbf{h} vectors using Algorithm \ref{algo:multi-gh-recovery} and finds out global optimal split node.

\end{itemize}

\begin{algorithm}[!ht] 
   \caption{SecureBoost-MO Split Info Recovery}
\label{algo:multi-gh-recovery}
     
    \textbf{Input:} \\
    $s$, a split info from the passive party; \\
    $l$, label numbers; \\
    $\eta_{c}$, number of class a ciphertext holds;
   
   \textbf{Output:} \\
    init $count = 0$; 
    
    init empty list $\textbf{g}, \textbf{h}$
    
    \begin{algorithmic}[1]
    \FOR{$e$ in $s.[[\textbf{vec}]]$}
    
        \STATE decrypt(e)
        
        \FOR{$i \leftarrow$  1 to $\eta_{c}$}
        
                \STATE $gh$ = $e \& (2^{b_{gh}} - 1)$, $e = e >> b_{gh}$
                
                \STATE $h = gh \& (2^{b_{h}} - 1)$, $g = gh >> b_{g}$
                
                \STATE $g = g - g_{off} \times s.sample\_count$
                
                \STATE $\textbf{g}.add(g), \textbf{h}.add(h)$
                
                \STATE $count += 1$
                
                \IF{$count == \eta_{c}$}
                   \STATE break
                \ENDIF
        \ENDFOR
    \ENDFOR
    
    $s.\textbf{g}, s.\textbf{h} = \textbf{g}, \textbf{h}$
    
    return $s$
    
\end{algorithmic}  
\end{algorithm}

\subsection{Other Optimization }
We also adopt two simple yet efficient engineering optimization methods in SecureBoost+ that are commonly used in the existing tree-boosting framework: GOSS~\cite{ke2017lightgbm} and Sparse Optimization.

\section{Experiments}
In this section, we conduct several experiments to verify the efficiency of SecureBoost+. 

\subsection{Setup}

\textbf{Baseline:} We use Secureboost provided in FATE v1.5 as the federated learning algorithm baseline and XGBoost as non-federated algorithm baseline.

\textbf{Environment:} Our experiments are conducted on two machines which are active and passive party respectively. Each machine has 16 cores and 32GB RAM. They are deployed in intranet with a network speed of 1GBps.

\textbf{Hyper-parameters:} For all experiments, key hyper-parameters settings are uniform: We set $tree\_depth=5$, $max\_bin\_num=32$ (In XGBoost the split method is 'hist') and $learning\_rate=0.3$. For encryption parameters, we use $Paillier$ as the encryption schema, the $key\_length=1024$. For the efficiency of conducting experiments, our baseline methods will run $tree\_num=25$ trees. 
In our framework SecureBoost+, we use goss subsample, set the $top\_rate=0.2$ and the $other\_rate=0.1$. In the Layered Tree Mode, set the ${h\_active=2}$ and ${h\_passive=3}$, and in the Mix Tree Mode every party is responsible for building $tree\_per\_party=1$ tree.

\textbf{Datasets:} We evaluate our framework on seven open data sets to test model performance and training speed. The detail of datasets is listed in the Table \ref{data set}.
The data sets contain large-instances and high-dimensional data. 
\begin{itemize}

 \item The first four data sets are for binary classification tasks:

    \textit{Give-credit}\footnote{https://github.com/FederatedAI/FATE/tree/master/examples/data}: A bank credit binary-classification data set. It contains 150, 000 instances and 10 features. It is extracted from UCI Credit Card data set and is a standard data set in FATE.
    
    \textit{Susy} and \textit{Higgs}: The data sets are to predict physics events. They have million scales of instances and are widely used in machine learning training speed evaluation \cite{chen2016xgboost}\cite{fu2019experimental}.
    
    \textit{Epsilon}: It is a binary data set from PASCAL Challenge 2008. It not only contains a large number of samples but also contains 2000 features. So It is an ideal data set to test the performance on the large-scale and high-dimensional modeling\cite{dorogush2018catboost}.

 \item  Three Multi-classification data sets are prepared for verifying the efficiency of SeucreBoost-MO:

    \textit{Sensorless}: It is for sensorless drive diagnosis.
    
    \textit{Covtype}: It is for predicting 7 types of forest covers.
    
    \textit{SVHN}: It is a street-view digits image classification task.
    
    We choose these three multi-classification data sets because they have more classes of labels while they have relatively more instances.
\end{itemize}
We vertically and equally divide every data set into two to make data sets for the active and passive party.

\begin{table*}[]
\centering
\caption{Data Set Details}
\label{data set}
\begin{tabular}{c|c|c|c|c|c|c}
\hline
\multicolumn{1}{c}{datasets} & \multicolumn{1}{c}{\# instance} & \multicolumn{1}{c}{\# features} & \multicolumn{1}{c}{\# active feature} & \multicolumn{1}{c}{\# passive features} & \multicolumn{1}{c}{\# labels} & \multicolumn{1}{c}{task type}\\
\hline
\hline
Give credit & 150,000     & 10          & 5                 & 5                & 2        & binary classification \\ \hline
Susy        & 5,000,000   & 18          & 4                 & 14               & 2        & binary classification \\ \hline
Higgs       & 11,000,000  & 28          & 13                & 15               & 2        & binary classification \\ \hline
Epsilon     & 400,000     & 2000        & 1000              & 1000             & 2        & binary classification \\ \hline
Sensorless  & 58,509      & 48          & 24                & 24               & 11       & multi-classification  \\ \hline
Covtype     & 581,012     & 54          & 27                & 27               & 7        & multi-classification  \\ \hline
SVHN     & 99,289     & 3072          & 1536                & 1536               & 10        & multi-classification  \\ \hline
\end{tabular}
\end{table*}

\subsection{Ciphertext Operation Optimization Evaluation}
In this section, we will conduct experiments to compare SecureBoost+ with the Baseline algorithms in the setting, which ciphertext operation optimization is applied in the SecureBoost+. SecureBoost+\_Ciphertext is defined as SecureBoost+ with ciphertext operation optimization.
\subsubsection{Training Time}
 We conduct experiments to compare the training time of SecureBoost+ with ciphertext operation optimization, and SecreuBoost on the four binary classification data sets. We only consider the time spent on tree building and ignore the time spent on non-tree building, such as time spent on data I/O, feature engineering and evaluation. For both SecureBoost+ and SecreuBoost, we both build 25 trees, and calculate the average time spent on each tree. The results are showed in Figure \ref{train_mechanism_time}. 
 We can see that the proposed SecureBoost+ obviously outperforms SecureBoost: In the four datasets, Secureboost+ are \textbf{6.63x, 6.03x, 7.33x, 22x} faster than SecureBoost. From the experimental results, we can find that with the increase of the number of instances and feature dimension, SecureBoost+ will have more obvious advantages.


\subsubsection{Model Performance}
In order to evaluate the impact of ciphertext operation optimization on the model performance, we compare the model performance of SecureBoost+, XGBoost and SecureBoost. We use AUC score as the model performance evaluation metric. The results are shown in Table \ref{auc_performance}. From the results, we can see that SecureBoost+ performs just as well as XGBoost and SecureBoost in this setting.


\subsection{Training Mechanism Optimization Evaluation}

In this section, we will conduct experiments to compare SecureBoost+ with the Baseline algorithms in the setting, which training mechanism optimizations are applied in the SecureBoost+.  It should be noted that in our experiment, when we enable Mix Tree Mode or Layered Tree Mode in the SecureBoost+,  we use ciphertext operation optimization in the SecureBoost+ by default.  SecureBoost+\_Mix is defined as SecureBoot+ with the mix tree mode, while SecureBoost+\_Layered is defined as SecureBoot+ with layered tree mode.

\subsubsection{Training Time}
\begin{figure}[htbp]
    \centering
    \includegraphics[width=0.48\textwidth]{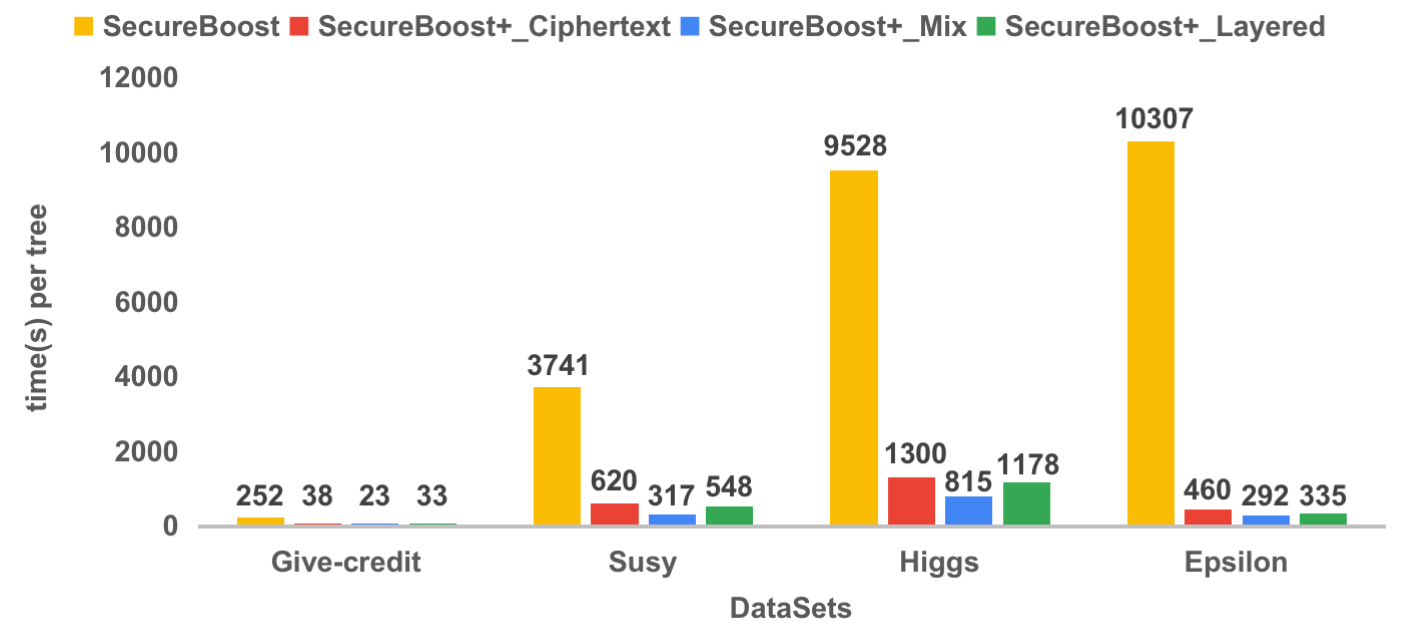}
    \caption{Comparison of Tree Building Time in Binary Classification} 
    \label{train_mechanism_time}
\end{figure}

We conduct experiments to compare the training time of SecureBoost+ with two training mechanism optimizations, SecreuBoost+ with ciphertext operation optimization, and SecureBoost on the four binary classification datasets. 
The results are showed in Figure \ref{train_mechanism_time}. It shows the average tree building time spent on each tree. It is no doubt that SecureBoost+ with the mix tree mode or layered tree mode will build tree faster than 
SecureBoost+ with ciphertext operation optimization only, and also faster than SecureBoost. This is because it skips global split-finding and inter-party communications with certain strategies. 
In the four datasets, SecureBoost+\_Mix are \textbf{1.65x, 1.96x, 1.60x, 1.58x} faster than SecureBoost+\_Ciphertext, while SecureBoost+\_Mix are \textbf{10.96x, 11.80x, 11.69x, 35.30x} faster than SecureBoost. 
In the four datasets, SecureBoost+\_Layered are \textbf{1.15x, 1.13x, 1.10x, 1.37x} faster than SecureBoost+\_Ciphertext., while SecureBoost+\_Layered are \textbf{7.64x, 6.83x, 8.09x, 30.77x} faster than SecureBoost. 

\subsubsection{Model Performance}
We will pay much attention to the model performance of SecureBoost+ with the mix tree mode and the layered tree mode since their strategies intentionally skip the global split finding on every tree node. The results are shown in Table \ref{auc_performance}. SecureBoost+\_Mix and SecureBoost+\_Layered have slight performance loss compared to XGBoost and SecureBoost+\_Ciphertext, but the losses are very small. According to our experimental analysis, the two mode sometimes require a few more trees to catch up with the performances of SecureBoost+\_Ciphertext. But from the result of \textit{Epsilon} data set, we believe that the two training mechanism can obviously accelerate training when the data set contains high dimensional features and features are balanced distributed.

\begin{table*}[]
\centering
\caption{Comparison of Model Performance in Binary Classification: AUC Metrics}
\label{auc_performance}
\begin{tabular}{c|c|c|c|c|c}
\hline
\multicolumn{1}{c}{Dataset} & \multicolumn{1}{c}{XGBoost} & \multicolumn{1}{c}{SecureBoost} & \multicolumn{1}{c}{SecureBoost+\_Ciphertext} & \multicolumn{1}{c}{SecureBoost+\_Mix} & \multicolumn{1}{c}{SecureBoost+\_Layered} \\
\hline
\hline
Give-credit      & 0.872  & 0.874    & 0.873   & 0.87  & 0.871   \\ \hline
Susy             & 0.864   & 0.873   & 0.873   & 0.869 & 0.87    \\ \hline
Higgs            & 0.808    & 0.806   & 0.8    & 0.795 & 0.796   \\ \hline
Epsilon          & 0.897   & 0.897   & 0.894   & 0.894 & 0.894   \\ \hline
\end{tabular}
\end{table*}

\subsection{SecureBoost-MO Optimization Evaluation}
In this section, we will conduct multi-classification task experiments to evaluate SecureBoost+ with SecureBoost-MO optimization.  We consider XGBoost and SecureBoost+\_Ciphertext as baseline. 

\subsubsection{Training Time}
\begin{figure}[!htbp]
    \centering
    \includegraphics[width=0.48\textwidth]{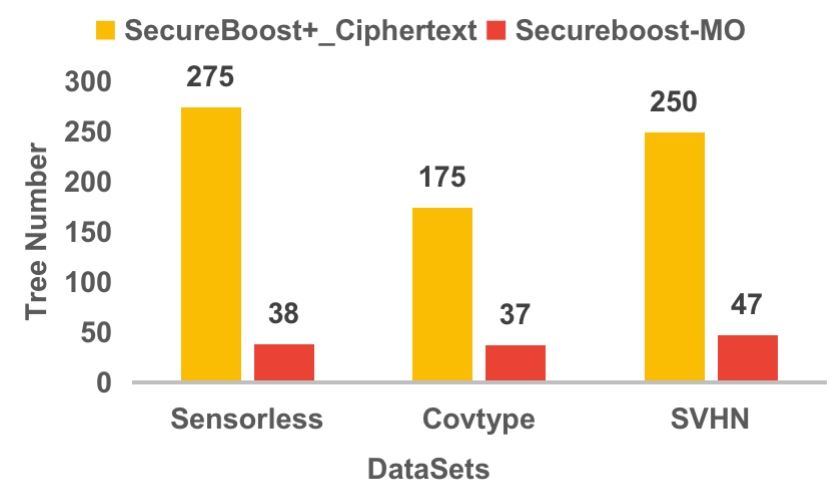}
    \caption{Tree number comparison}
    \label{tree-num}
\end{figure}

 As shown in Figure \ref{tree-num}, in the multi-classification training task with 25 epoches, SecureBoost+\_Ciphertext needs to build 275, 175 and 250 trees respectively in the \textit{Sensorless}, \textit{Covtype}, \textit{SVHN} datasets, while SecureBoost-MO only needs to build 38, 37, 47 trees respectively. The tree numbers of SecureBoost-MO is far less than SecureBoost+\_Ciphertext. 
 
\begin{figure}[htbp]
    \centering
    \includegraphics[width=0.48\textwidth]{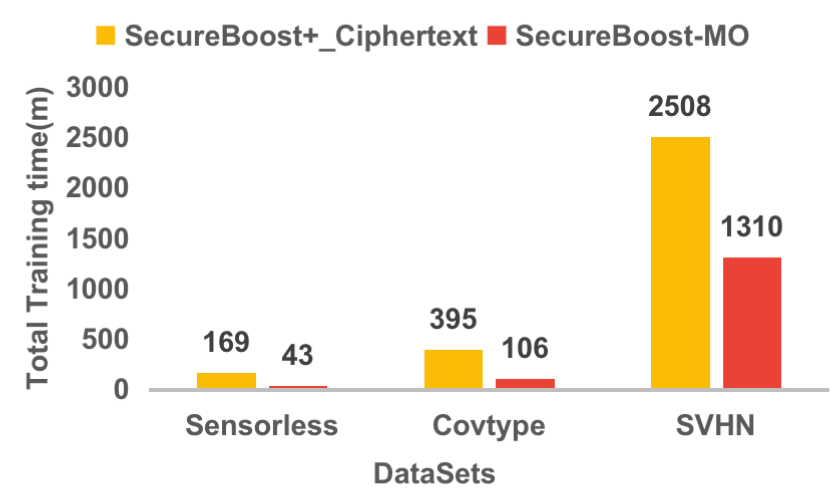}
    \caption{Comparison of Tree Building Time in Multi-Classification} 
    \label{mo-time}
\end{figure}

We conduct experiments to compare the training time of SecureBoost-MO with SecureBoost+\_Ciphertext on the three multi-classification datasets. It should be noted that in this experiment, the time is the total training time, not the training time for each tree. In the SecureBoost+\_Ciphertext, we run 25 epoch, which has 275 trees in the Sensorless dataset, while in the SecureBoost-MO, we need to build 38 trees to achieve the similar model performance.
The results are showed in Figure \ref{mo-time}.  We can see that SecureBoost-MO obviously outperforms SecureBoost+\_Ciphertext: In the three datasets, SecureBoost-MO are \textbf{3.92x, 3.72x, 1.57x} faster than SecureBoost+\_Ciphertext.

\subsubsection{Model Performance}
  We use accuracy scores as the model performance evaluation metric in the multi-classification tasks. The results are shown in Table \ref{multi-acc}. The results show that SecureBoost+\_Ciphertext has relatively better accuracy scores than XGBoost in some datasets, and SecureBoost-MO performs as well as SecureBoost+\_Ciphertext.

\begin{table}[]
\centering
\caption{Comparison of Model Performance in Multi-Classification with accuracy as the metric. SecureBoost+\_CT denotes SecureBoost+\_Ciphertext.}
\label{multi-acc}
\begin{tabular}{c|c|c|c}
\hline
\multicolumn{1}{c}{Data set} & \multicolumn{1}{c}{XGBoost} & \multicolumn{1}{c}{SecureBoost+\_CT} &  \multicolumn{1}{c}{SecureBoost-MO}\\
\hline
\hline
Sensorless                & 0.999   & 0.992  & 0.994      \\ \hline
Covtype                       & 0.78    & 0.806  & 0.807      \\ \hline
SVHN                      & 0.686   & 0.686  & 0.685      \\ \hline
\end{tabular}
\end{table}

\section{Conclusion}
In this work, we proposed SecureBoost+, a large-scale and high-performance vertical federated gradient boosting decision tree framework. Based on Secureboost, we propose several optimization schemes, including Ciphertext Operation Optimization, Training Mechanism Optimization and SecureBoost-MO Optimization. These optimization schemes cover binary and multi-classification task scenarios. The experimental results show that the speed of SecureBoost+ is significantly faster than that of SecureBoost, and the accuracy of SecureBoost+ is comparable to that of SecureBoost and XGBoost. With the increase of the number of instances and feature dimension, SecureBoost+ will have more obvious advantages.


\bibliographystyle{ACM-Reference-Format}
\bibliography{refsecureboost}

\end{document}